%% file: main.tex
\documentclass[10pt,twocolumn,letterpaper]{article}

\usepackage[pagenumbers]{cvpr} 

\usepackage{graphicx}
\usepackage{amsmath}
\usepackage{amssymb}
\usepackage{booktabs}
\usepackage{tabularx}
\usepackage{multirow}
\usepackage{multicol}
\usepackage{multicol}
\usepackage{bm}
\usepackage[accsupp]{axessibility}

\usepackage[pagebackref,breaklinks,colorlinks]{hyperref}

\usepackage[capitalize]{cleveref}
\crefname{section}{Sec.}{Secs.}
\Crefname{section}{Section}{Sections}
\Crefname{table}{Table}{Tables}
\crefname{table}{Tab.}{Tabs.}

\def\cvprPaperID{3664} 
\def\confName{CVPR}
\def\confYear{2023}

\input{preamble}
\begin{document}

\renewcommand{\thefootnote}{\fnsymbol{footnote}}

\title{CAMS: CAnonicalized Manipulation Spaces for\\Category-Level Functional Hand-Object Manipulation Synthesis}


\author{
Juntian Zheng\textsuperscript{\textasteriskcentered,1,3}~~~
Qingyuan Zheng\textsuperscript{\textasteriskcentered,3}~~~
Lixing Fang\textsuperscript{\textasteriskcentered,1,3}~~~
Yun Liu\textsuperscript{1}~~~
Li Yi\textsuperscript{\textdagger,1,2,3}
\smallskip\\
\textsuperscript{1}IIIS, Tsinghua University~~~~~~
\textsuperscript{2}Shanghai Artificial Intelligence Laboratory\\
\textsuperscript{3}Shanghai Qi Zhi Institute
}


\twocolumn[{
\renewcommand\twocolumn[1][]{#1}
\maketitle
\begin{center}
    \captionsetup{type=figure}
    \vspace{-20pt}
    \includegraphics[width=0.95\textwidth]{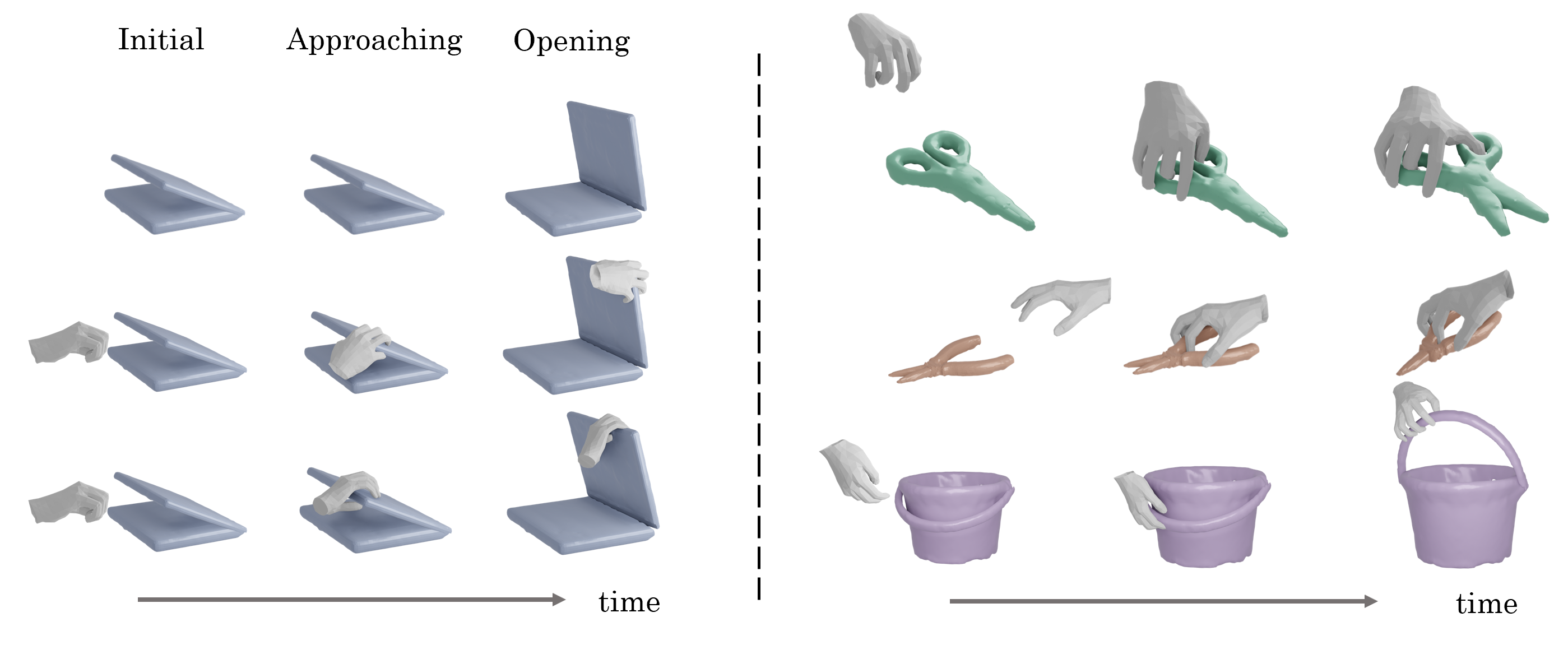}
    \vspace{-20pt}
    \caption{Given a sequence of manipulation goals, our method can generate realistic and diverse functional manipulation motions consistent with the goal sequence. The motions are expressed in snapshots at several keyframes. On the left side, we show a goal sequence of opening a laptop and two different manipulation patterns that can be generated by our method. On the right side, we show our generation results of three manipulation tasks corresponding to different object categories.}
    \label{fig:Head}
\end{center}
}]

\footnotetext[1]{Equal contribution with the order determined by rolling dice.}
\footnotetext[2]{Corresponding author.}

\begin{abstract}
\input{tex/abstract.tex}
\end{abstract}

\section{Introduction}
\input{tex/introduction.tex}

\section{Related Work}
\input{tex/related_work.tex}
\section{Problem Formulation and Notations}
\label{sec:overview}
\input{tex/overview}

\section{Method}

\input{tex/method.tex}


\section{Experiments}
\input{tex/exp.tex}

\section{Conclusion}
\input{tex/conclusion}

{\small
\bibliographystyle{ieee_fullname}
\bibliography{egbib}
}

\end{document}

%% file: preamble.tex
\newcommand{\figref}[1]{Fig.~\ref{#1}}
\newcommand{\tabref}[1]{Tab.~\ref{#1}}

%% file: tex/abstract.tex
\vspace{-10pt}
In this work, we focus on a novel task of category-level functional hand-object manipulation synthesis covering both rigid and articulated object categories. Given an object geometry, an initial human hand pose as well as a sparse control sequence of object poses, our goal is to generate a physically reasonable hand-object manipulation sequence that performs like human beings. To address such a challenge, we first design CAnonicalized Manipulation Spaces (CAMS), a two-level space hierarchy that canonicalizes the hand poses in an object-centric and contact-centric view. Benefiting from the representation capability of CAMS, we then present a two-stage framework for synthesizing human-like manipulation animations. Our framework achieves state-of-the-art performance for both rigid and articulated categories with impressive visual effects. Codes and video results can be found at our project homepage: \url{https://cams-hoi.github.io/}.

%% file: tex/introduction.tex
Human conducts hand-object manipulation (HOM) for certain functional purposes commonly in daily life, \eg opening a laptop and using scissors to cut. Understanding how such manipulation happens and being able to synthesize realistic hand-object manipulation has naturally become a key problem in computer vision. A generative model that can synthesize human-like functional hand-object manipulation plays an essential role in various applications, including video games, virtual reality, dexterous robotic manipulation, and human-robot interaction.

This problem has only been studied with a very limited scope previously. Most existing works focus on the synthesis of a static grasp either with \cite{ContactGrasp} or without \cite{grasptta} a functional goal. Recently, there have been works started focusing on dynamic manipulation synthesis~\cite{zhang2021manipnet,Dgrasp}. However, these works restrict their scope to rigid objects and do not consider the fact that functional manipulation might change the object geometry as well, such as in opening a laptop by hand. Moreover, these works usually require a strong input, including hand and object trajectories or a grasp reference, limiting their application scenarios.

To expand the scope of HOM synthesis, we propose a new task of category-level functional hand-object manipulation synthesis. Given a 3D shape from a known category as well as a sequence of functional goals, our task is to synthesize human-like and physically realistic hand-object manipulation to sequentially realize the goals as shown in Figure~\ref{fig:Head}. Besides rigid objects, we also consider articulated objects, which support richer manipulations than a simple move. We represent a functional goal as a 6D pose for each rigid part of the object. We emphasize category-level for generalization to unseen geometry and for more human-like manipulations revealing the underlying semantics.

In this work, we choose to tackle the above task with a learning approach. We can learn from human demonstrations for HOM synthesis thanks to the recent effort in capturing category-level human-object manipulation dataset~\cite{hoi4d}. The key challenges lie in three aspects. First, a synthesizer needs to generalize to a diverse set of geometry with complex kinematic structures. Second, humans can interact with an object in diverse ways. Faithfully capturing such distribution and synthesizing in a similar manner is difficult. Third, physically realistic synthesis requires understanding the complex dynamics between the hand and the object. Such understanding makes sure that the synthesized hand motion indeed drives the object state change without violating basic physical rules.

To address the above challenges, we choose to generate object motion through motion planning and learn a neural synthesizer to generate dynamic hand motion accordingly.
Our key idea is to canonicalize the hand pose in an object-centric and contact-centric view so that the neural synthesizer only needs to capture a compact distribution. This idea comes from the following key observations. During functional hand-object manipulation, human hands usually possess a strong preference for the contact regions, and such preference is highly correlated to the object geometry, \eg hand grasping the display edge while opening a laptop. From the contact point's perspective, the finger pose also lies in a low-dimensional space. Representing hand poses from an object-centric view as a set of contact points and from a contact-centric view as a set of local finger embeddings could greatly reduce the learning complexity.

Specifically, given an input object plus several functional goals, we first interpolate per-part object poses between every adjacent functional goal, resulting in an object motion trajectory. Then we take a two-stage method to synthesize the corresponding hand motion. In the first stage, we introduce CAnonicalized Manipulation Spaces (CAMS) to plan the hand motion. CAMS is defined as a two-level space hierarchy. At the root level, all corresponding parts from the category of interest are scale-normalized and consistently oriented so that the distribution of possible contact points becomes concentrated. At the leaf level, each contact point would define a local frame. This local frame would simplify the distribution of the corresponding finger pose. With CAMS, we could represent a hand pose as an object-centric and contact-centric CAMS embedding. At the core of our method is a conditional variation auto-encoder, which learns to predict a CAMS embedding sequence given an object motion trajectory. In the second stage, we introduce a contact- and penetration-aware motion synthesizer to further synthesize an object motion-compatible hand motion given the CAMS embedding sequence.

To summarize, our main contributions include: i) A new task of functional category-level hand-object manipulation synthesis. ii) CAMS, a hierarchy of spaces canonicalizing category-level HOM enabling manipulation synthesis for unseen objects. iii) A two-stage motion synthesis method to synthesize human-like and physically realistic HOM. iv) State-of-the-art HOM synthesis results for both articulated and rigid object categories.



\begin{figure*}[!htb]
\begin{center}
   \includegraphics[width=2\columnwidth]{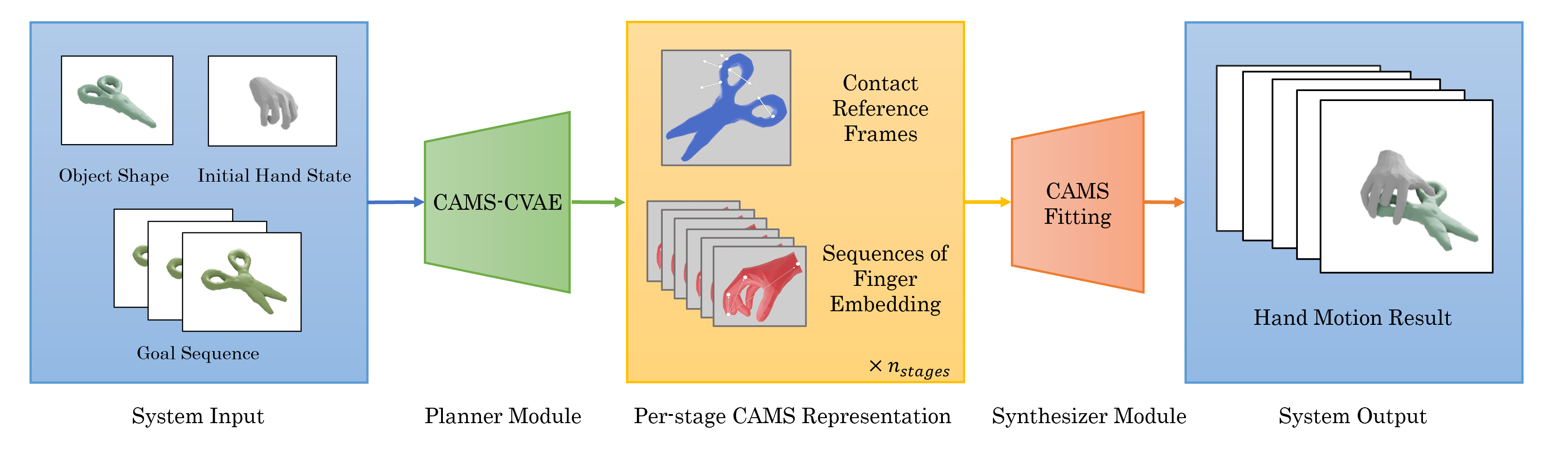}
\end{center}
   \vspace{-0.7cm}
   \caption{\textbf{System Overview.} Our framework mainly consists of a CVAE-based planner module and an optimization-based synthesizer module. Given the generation condition as the input, the planner first generates a per-stage CAMS representation containing contact reference frames and sequences of finger embedding. Then the synthesizer optimizes the whole manipulation animation based on the CAMS embedding.}
   \vspace{-0.3cm}
\label{fig:framework}
\end{figure*}

%% file: tex/related_work.tex
\label{sec:relatedwork}

\subsection{Human Motion Synthesis}

Human motion synthesis, including motion prediction, interpolation, and completion, has attracted many interests these years \cite{humanmotion4, humanmotion1, zhang2021mojo, cai2021unifiedhumancvae, humanmotion2, humanmotion3, transcvae, actionhumanmotion, henter2020moglow, longtermsceneawarecvae, awarehuman, couch, li2021ai, avatarclip, huang2023diffusion, lee2023locomotion, saga, zhang2022wanderings}. Conditional variational autoencoder\cite{cvae} (CVAE) was widely used\cite{zhang2021mojo, cai2021unifiedhumancvae, actionhumanmotion, transcvae} for its generalizability across various scenes and human motions. 
These works have achieved great success in human motion synthesis, while their potential in HOM synthesis was overlooked. In particular, \cite{longtermsceneawarecvae, awarehuman, couch} focused on modeling human-scene interaction when generating human motion. Such scene-aware or context-aware methods might also suit HOM synthesis.

\subsection{Physics-Based Object Manipulation Synthesis}

Generating high-quality human grasps remains challenging due to the complex geometry and complicated skeletal constraints. Physics-based methods\cite{samplebased1, samplebased2, physic1, ibs, Dgrasp, chopstick} were favored for in-hand manipulation synthesis since the generated motion was physically plausible. IBS\cite{ibs} presented a novel representation of hand-object interaction and leveraged reinforcement learning (RL) methods with execution success, and geometric measure \cite{219918} rewards to generate successful grasping motion. D-Grasp\cite{Dgrasp} developed its grasping policy based on the physical attributes of hand and object, including angles and velocities. Yang \etal \cite{chopstick} concentrated on using chopsticks in diverse gripping styles, and it solved this rather difficult task by first optimizing physically valid gripping poses with predefined gripping styles and then utilizing the carefully designed hand-controlled policies to synthesize manipulation.

\subsection{Data-Driven Object Manipulation Synthesis}

Besides physics-based methods, a line of data-driven approaches\cite{graspingfield, grasptta, CPFChen_2022, static1, static2, static3, static4, static5, zhang2021manipnet, GRAB:2020} could generate manipulation in a more natural and human-like manner and generalize to novel object instances. Most of the previous data-driven works focused on reconstruction and synthesis for the static grasp\cite{graspingfield, grasptta, CPFChen_2022, static1, static2, static3, static4, static5}. Grasping Field\cite{graspingfield} and CPF\cite{CPFChen_2022} reconstructed a static grasping field in 3D space for hand-object interaction based on an RGB image. GraspTTA\cite{grasptta} predicted a static joint-angle configuration of the grasping hand from a given object point cloud and contributed a Test Time Adaptation (TTA) strategy for helping the method generalize to novel objects.
Going beyond static grasping synthesis,  ManipNet\cite{zhang2021manipnet} proposed several geometric sensors and managed to generate long-term complex manipulation sequences. TOCH\cite{TOCH} achieves a data-driven approach of dynamic motion refinement in hand object motion synthesis.

%% file: tex/overview.tex
In this paper, we focus on synthesizing functional manipulation for a specific task goal defined on a known category of articulated or rigid objects (\eg opening a laptop). The input of our system can be split into three parts:

\begin{itemize}
    \item A \textbf{novel object instance} from a known object category $\mathcal{C}$. The object instance is described by triangular object part meshes $\{\mathbf{M}_k\}_{k=1}^N$, where $N>1$ indicates an articulated object and $N=1$ indicates a rigid object. We assume that the number of rigid parts $N$ for each object category is known and constant.
    \item A \textbf{goal sequence} $\mathbf{G}=\{(\mathbf{S}^j\to \mathbf{S}^{j+1}, \mathbf{t}^j)\}_{j=0}^M$ that defines the goal of a manipulation task by the movement of object's parts, in which $t^j$ denotes the lasting time of the $j$-th stage, and $\mathbf{S}^j=\{\mathbf{S}^j_k\in SE(3)\}_{k=1}^N$ denotes the 6D poses of each rigid part at the $j$-th stage transition point. The whole manipulation procedure is subdivided into multiple stages according to the object's state of movement (\eg A goal sequence of the \textit{opening} task for laptops consists of two stages: \textit{approaching} and \textit{opening}, as shown in Figure~\ref{fig:Head}). 
    \item An \textbf{initial hand pose} $\mathbf{H}_0\in\mathbb R^{51}$ represented by the $48$-dimensional MANO\cite{mano} pose parameters and a 3D wrist position.
\end{itemize}

The goal is to generate a sequence of human-like hand motions that is represented by a hand pose sequence $\{\theta_t\in\mathbb{R}^{51}\}_{t=1}^T$. Such a sequence of motion should be consistent with the given goal sequence of the manipulated object. To make this generation task reasonable, the given goal sequence should conform to the object's articulation constraint and follow its functionality.



%% file: tex/method.tex
\label{sec:method}

Figure~\ref{fig:framework} shows the framework of our system. We first introduce \textbf{CAnonicalized Manipulation Spaces} (\textbf{CAMS}), a two-level space hierarchy that allows representing dynamic hand motions in an object-centric and contact-centric view (see Section~\ref{sec:cams}).
By embedding hand motions in CAMS, we obtain a more compact description for dynamic manipulations, which is more friendly to learning.

To learn to synthesize human-like manipulations, we propose a two-stage framework consisting of a CVAE-based \textbf{planner} module (see Section~\ref{sec:planner}) and an optimization-based \textbf{synthesizer} module (see Section~\ref{sec:synthesizer}). Taking the object shape $\mathbf{O}$, the manipulation goal $\mathbf{G}$ and the initial hand configuration $\mathbf{H}_0$ as input, we first divide the whole manipulation process into several motion stages by the sequential goals. Then, the planner module generates stage-wise \textbf{contact targets} and time-continuous \textbf{finger embeddings} as guidance for HOM synthesis. Finally, the synthesizer module takes the generated contact targets and hand embeddings as inputs and leverages an optimization strategy to produce a human-like and physically realistic HOM animation.

\begin{figure}
    \centering
    \vspace{-5pt}
    \includegraphics[width=\linewidth]{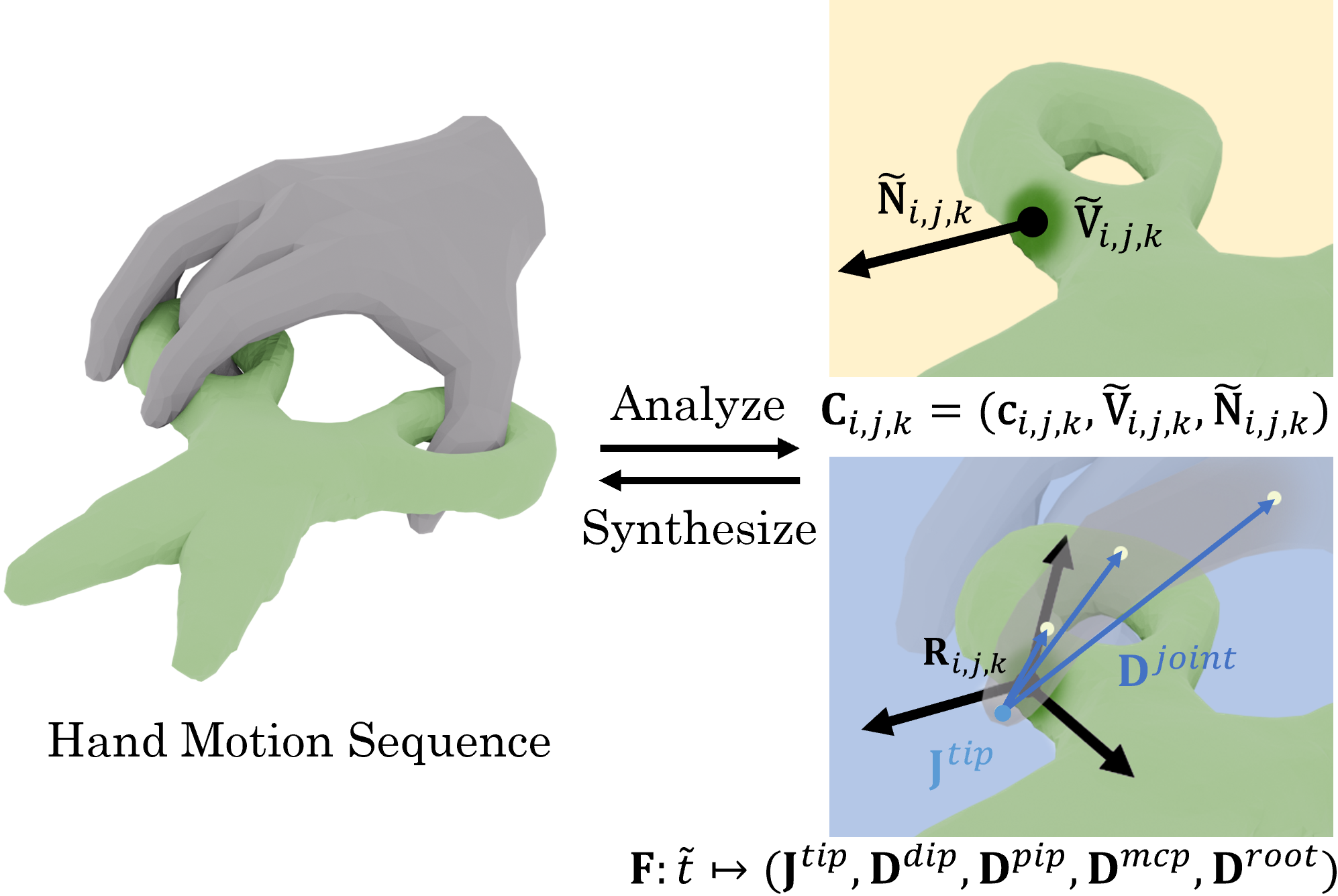}
    \vspace{-10pt}
    \caption{\textbf{CAMS Representation of Hand Motion.} We present Canonicalized Manipulation Spaces, a novel representation for hand-object interaction, to express the motion under an object-centric and contact-centric view. The top right figure demonstrates discrete contact targets defined at each stage transition point. The bottom right figure shows the time-continuous finger embedding in one frame. See Section~\ref{sec:cams} for details.}
    \label{fig:CAMS}
    \vspace{-20pt}
\end{figure}

\subsection{CAnonicalized Manipulation Spaces (CAMS)}
\label{sec:cams}

It is challenging to model the space of hand manipulation from the given object shape due to the shape variety within a category and the huge diversity of human manipulation styles. For two object instances that have a non-negligible geometry difference, even if we apply the same manipulation style on them (\eg same finger placement on two pairs of scissors with different sizes), the resulting hand pose can also be significantly different. Hence it is difficult for previous approaches that directly learn the MANO parameters to generalize to novel object instances, especially when object shapes are similar during training. 
To address this challenge, we propose to express the motion of each finger in a canonical reference frame centered at a contact point on the object, thus associating such motion to the local contact region rather than the whole object shape.

Based on this motion expression, we introduce CAnonicalized Manipulation Spaces with two-level canonicalization for manipulation representation. At the root level, the \textit{canonicalized contact targets} (see Section~\ref{subsec:CR}) describe the discrete contact information. At the leaf level, the \textit{canonicalized finger embedding} (see Section~\ref{subsec:FE}) transforms finger motion from global space into local reference frames defined on the contact targets.



\begin{figure*}[!h]
\begin{center}
   \includegraphics[width=2\columnwidth]{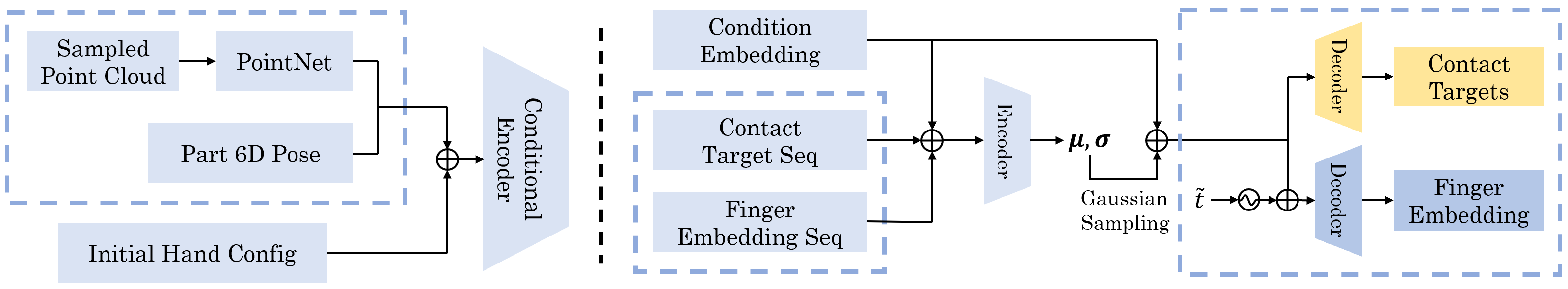}
\end{center}
   \vspace{-0.5cm}
   \caption{\textbf{Network Architecture of CAMS-CVAE.} On the left side, we show the design of the condition encoder, in which the system input is encoded into a condition embedding for CVAE. The components in the dotted box are repeated for each rigid part of the object. On the right side, we show the architecture of the whole CVAE. The components in the dotted boxes are repeated for each tuple $(i,j,k)$.}
   \vspace{-0.3cm}
\label{fig:network}
\end{figure*}

\subsubsection{Canonicalized Contact Targets}
\label{subsec:CR}

At the root level of CAMS, the canonicalized contact targets describe the contact information between each finger and the object. The contact targets are first used to define the contact-centric reference frames for finger embeddings (Section~\ref{subsec:FE}), and then allow the contact optimization of the synthesizer to achieve grasps with accurate contacts (Section~\ref{sec:synthesizer}).

Formally, we define the contact targets as a sequence

\vspace{-10pt}
$$\mathbf{C}=\{(\mathbf{c}_{i,j,k},\mathbf{\Tilde{V}}_{i,j,k}, \mathbf{\Tilde N}_{i,j,k})\}_{i,j,k},$$

in which $1\le i\le 5$ indicates the index of finger, $0\le j\le M$ indicates the index of stage transition point and $1\le k\le N$ indicates the index of object's part. $\mathbf c_{i,j,k}\in[0,1]$ are binary flags of whether the finger $i$ is in contact with the object part $k$ at stage transition point $j$, and $\mathbf{\Tilde V}_{i,j,k}\in\mathbb R^3,\mathbf{\Tilde N}_{i,j,k}\in\mathbb R^3$ are canonicalized positions and normal directions of the corresponding contact point. We follow NPCS~\cite{nocs, li2020category} to normalize all the rigid parts into a normalized coordinate space within the unit cube (i.e., $x, y, z \in [0,1]$), aligned with a category-level canonical orientation. Contact positions $\mathbf{\Tilde V}_{i,j,k}$ as well as the surface normals $\mathbf{\Tilde N}_{i,j,k}$ are defined in the normalized space, in order to make the representation more compactly distributed and easier to learn.

Our generative model learns the distribution of discrete contact targets conditioned on the object shape and task configuration. The benifit of learning surface normal $\mathbf{\Tilde{N}}_{i,j,k}$ together with position $\mathbf{\Tilde{V}}_{i,j,k}$ is two fold. First, for an unseen object with complex shapes (\eg scissors), due to the limitation of network capacity and training data diversity, we cannot guarantee that the predicted contact positions are always exactly on the object surface. When projecting the contact point onto the surface, the predicted normal directions help filter out the incorrect surface parts near the predicted point. Second, the difference in normal directions among different finger placement styles is more significant. It makes it easier for the VAE-based model to distinguish the discrete manipulation styles if normal directions are regarded as reconstruction targets. 


At the training stage, the ground truth contact targets are obtained by applying a contact analysis to training data. At the inference stage, given $\mathbf {\Tilde V}_{i,j,k}$ and $\mathbf{\Tilde N}_{i,j,k}$ with $\mathbf c_{i,j,k}=1$, the actual contact point $\mathbf P_{i,j,k}$ on the object surface is determined by a matching process on the object surface. For more details, please refer to our supplementary material.

\subsubsection{Canonicalized Finger Embeddings}
\label{subsec:FE}

Given contact points $\mathbf P_{i,j,k}$ from contact targets, we can now build contact-centric spaces for finger embedding. We denote $\mathbf R_{i,j,k}$ as the contact reference frame centered at $\mathbf P_{i,j,k}$, with orientation aligned to the corresponding rigid part of the object (as shown in \ref{fig:CAMS}). For a static hand pose represented by MANO parameters $\theta$ and $\beta$, we first calculate the corresponding MANO joint coordinates $\mathbf J^{tip},\mathbf J^{dip},\mathbf J^{pip},\mathbf J^{mcp}$ of finger $i$ and $\mathbf J^{root}$ for the hand wrist under reference frame $\mathbf{R}_{i,j,k}$. Based on joint positions, we further compute the normalized directions
\vspace{-5pt}
$$
\mathbf{D}^{joint}=\frac{\mathbf{J}^{joint}-\mathbf J^{tip}}{\|\mathbf{J}^{joint}-\mathbf J^{tip}\|},joint\in\{dip,pip,mcp,root\}
$$
from the fingertip to every other joint. The resulting finger embedding for the static hand pose is $\mathbf F=(\mathbf J^{tip},\mathbf D^{dip},\mathbf D^{pip},\mathbf D^{mcp},\mathbf D^{root})\in\mathbb R^{15}$. Similar to contact targets, we aim to enlarge the discrepancy between different modes by using directions as learning targets. Also, directions between joints are invariant under global translation relative to the contact center, and therefore data noise caused by the contact center's fluctuation can be reduced. The tip joint is selected as the reference joint since it's typically close to the contact position in most of the manipulation styles.

To handle dynamic hand motion synthesis, we further extend the static finger embedding to \textit{continuous-time sequences}. More specifically, for a finger interacting with the object at a stage in time period $[t_0,t_1]$, the continuous-time finger embedding sequence is a continuous function $\mathbf f:[0,1]\to\mathbb R^{15}$ that maps the normalized time $\Tilde t=(t-t_0)/(t_1-t_0)$ to a finger embedding $\mathbf F\in\mathbb F^{15}$. Inspired by recent works\cite{nerf, nerfies, d-nerf} of implicit neural representation, the continuous mapping is learned by a neural network which maps the \textit{temporal encoding} $\mathcal T\in\mathbb R^{12}$ defined as
$$\mathcal T(\Tilde t)=\{(\sin2^k\pi \Tilde t,\cos2^k\pi \Tilde t)\}_{k=0}^5$$
together with a latent code to $\mathbf{F}\in\mathbb R^{15}$ (see Section~\ref{sec:planner}).

\subsection{CAMS-CVAE: the Motion Planner}
\label{sec:planner}

Given an object instance with task configuration, we present CAMS-CVAE, a CVAE-based generative motion planner for generating CAMS sequences, including contact targets and continuous-time finger embeddings. An overview of the model architecture is shown in Figure~\ref{fig:network}.

\textbf{Condition Encoding} As mentioned earlier, our model takes several inputs as generating conditions, including the object part meshes $\{\mathbf M_k\}_{k=1}^N$, the goal sequence $\mathbf G$ and the initial hand configuration $\mathbf H_0$. We encode this condition information into a vector using a multi-head condition encoder module, with each head corresponding to an object's rigid part. For each part, we first sample $2000$ points on the mesh surface with normal directions, and a PointNet structure is used to embed the point cloud information $\mathbf O\in\mathbb R^{2000\times6}$ into a shape feature $\bm z_{O}\in\mathbb R^{32}$. The shape feature is then concatenated with the part's 6D-pose sequence $\mathbf S\in\mathbb R^{6\times(M+1)}$, and all part's information is concatenated together with the initial hand pose $\mathbf H_0\in\mathbb R^{51}$. Finally, an MLP structure encodes the concatenated vector into the condition embedding $\bm z_C\in\mathbb R^{32}$.

\textbf{Motion Encoding} In the encoder part of the CVAE structure, we encode the hand motion represented by CAMS sequences into a diagonal-covariance Gaussian distribution in a $64$-dimensional latent space. We first concatenate the whole sequence of contact targets into a vector $\mathbf C\in\mathbb R^{7\times5\times N\times(M+1)}$ (as introduced in Section~\ref{subsec:CR}). In each stage, we evenly spaced sample $10$ timestamps $\{0,1/9,2/9,\cdots,1\}$ in the normalized time range and calculated the corresponding finger embeddings (as introduced in Section~\ref{subsec:FE}) under contact targets at the stage's two end-points. Besides finger embeddings, we also extract two binary flags $\textbf f_c,\textbf f_n$ (see Section~\ref{sec:synthesizer}) used in the synthesizer module at each sampled timestamp for each tuple $(i,j,k)$. For non-existing contact targets with $\mathbf c_{i,j,k}=0$, the corresponding values are filled with zero. All these values are concatenated are encoded by an MLP to predicted Gaussian parameters $\bm\mu\in\mathbb R^{64},\bm\sigma\in\mathbb R^{64}$.

\textbf{Motion Decoding} After sampling latent code $\bm z\in\mathbb R^{64}$ from either the predicted distribution $\mathcal N(\bm\mu,\bm\sigma^2)$ or standard Gaussian $\mathcal N(\mathbf 0,\mathbf I)$, we concatenate it with the generating condition $\bm z_C$, and a two-branch decoder is used to convert them into CAMS representation of desired hand motion. The \textit{discrete branch} is a multi-head MLP that outputs discrete contact targets $(\mathbf{c}_{i,j,k},\mathbf{\Tilde{V}}_{i,j,k}, \mathbf{\Tilde N}_{i,j,k})\in\mathbb R^7$ introduced in Section~\ref{subsec:CR}, with each head corresponding to a tuple $(i,j,k)$ indicating the index of the finger, stage transition point, and object's part. The \textit{continuous-time branch} is another multi-head MLP that takes the temporal encoding $\mathcal{T}(\Tilde{t})\in\mathbb R^{12}$ as extra input and outputs the corresponding finger embeddings $\mathbf F_1,\mathbf F_2\in\mathbb R^{15}$ relative to contact targets at both end-points of the stage. The continuous-time branch also predicts the two binary flags $\textbf f_c,\textbf f_n$ for the synthesizer.


\textbf{Training Loss}
We use several losses to train CAMS-CVAE. The first loss is a Binary Cross Entropy (BCE) loss $\mathcal{L}_{flag}$ on the contact target flags $\mathbf c_{i,j,k}$ and per-frame flags $\mathbf f_c,\mathbf f_n$ for synthesizer, between predicted values and the ground truth.
We then calculate the $L_2$ distance losses $\mathcal L_{pos}$ of $\mathbf{\Tilde{V}}_{i,j,k}$ and $\mathcal L_{dir}$ for $\mathbf{\Tilde N}_{i,j,k}$, between predicted values and the ground truth. These two losses are computed only for $(i,j,k)$ with the ground truth contact flag $\mathbf c_{i,j,k}=1$.
Similarly, we also compute the $L_2$ losses $\mathcal{L}_{tip}$ for predicted $\mathbf J^{tip}$ and $\mathcal{L}_{vec}$ for predicted $\mathbf D^{dip},\mathbf D^{pip},\mathbf D^{mcp},\mathbf D^{root}$.
Finally, we use the KL-Divergence $\mathcal{L}_{KLD}$ to constrain the latent distribution $\mathcal N(\bm\mu,\bm\sigma^2)$ to be close to the standard Gaussian distribution following \cite{vae}.
The total loss is a weighted summation of all six loss terms:
\begin{equation}
\begin{split}
    \mathcal{L} & = \lambda_{flag} \cdot \mathcal{L}_{flag} + \lambda_{pos} \cdot \mathcal{L}_{pos} + \lambda_{dir} \cdot \mathcal{L}_{dir} \\ & + \lambda_{tip} \cdot \mathcal{L}_{tip} + \lambda_{vec} \cdot \mathcal{L}_{vec} + \lambda_{kld} \cdot \mathcal{L}_{KLD}.
\end{split}
\end{equation}

For a detailed calculation of the loss terms, please refer to our released code.

\subsection{Optimization-Based Motion Synthesizer}
\label{sec:synthesizer}

Once a CAMS embedding sequence has been generated from the planner, our optimization-based synthesizer subsequently produces a complete HOM sequence. We simply use Bézier curve-based interpolation to generate the object trajectory and thus focus on synthesizing a hand trajectory that satisfies our goal (Section~\ref{sec:overview}).
Given the object shape and trajectory, as well as a CAMS embedding sequence, our synthesizer adopts a two-stage optimization method that first optimizes the MANO pose parameters to best fit the CAMS finger embedding (see Section~\ref{subsec:fitting_finger_embedding}) and then optimizes the contact effect to improve physical plausibility (see Section~\ref{subsec:optimizing_contact_and_penetration}).
In each frame, the synthesizer reads a binary flag $\mathbf f_n$ indicating whether the finger is near enough (10cm) that the finger embedding will be used to guide the generated motion, and a binary flag $\mathbf f_c$ indicating whether there is a contact between the finger and object.


\subsubsection{Fitting Finger Embedding}
\label{subsec:fitting_finger_embedding}

Given the predicted contact targets $\mathbf{C}$, the bidirectional finger embedding $\mathbf{F}_1,\mathbf{F}_2$ and the binary flags $\mathbf{f}_c,\mathbf{f}_n$, we optimize the MANO parameters of hand pose and transition $\theta=\{\theta_t\}_{t=1}^{T}$ by minimizing:
\vspace{-20pt}

\begin{equation} \label{eq:fit_loss}
\begin{split}
    \mathcal{L}(\theta) = & \lambda_{\mathrm{tip}}\sum_{t=1}^{T}\mathcal{L}_{\mathrm{tip}}(\theta_t)+\lambda_{\mathrm{joint}}\sum_{t=1}^{T}\mathcal{L}_{\mathrm{joint}}(\theta_t) \\
    &+ \lambda_{\mathrm{smooth}}\mathcal{L}_{\mathrm{smooth}}(\theta).
\end{split}
\end{equation}

The first term of Eq.\eqref{eq:fit_loss} is the tip transition loss, which constrains the tip position $\mathbf{J}^{tip}$ of the finger in the canonicalized finger embedding. The second term of Eq.\eqref{eq:fit_loss} is the joint orientation loss, which is used to optimize the direction vectors of four subsequent joints $\mathbf{D}^{dip}$, $\mathbf{D}^{pip}$, $\mathbf{D}^{mcp}$, $\mathbf{D}^{root}$ of the finger in the canonicalized finger embedding. And the last term of \eqref{eq:fit_loss} is a smoothness loss for improving temporal continuity.

\subsubsection{Optimizing Contact and Penetration}
\label{subsec:optimizing_contact_and_penetration}

After fitting MANO parameters $\theta$ by finger embedding, we leverage another optimization-based method to handle the penetration and inaccurate contact issues of the hand pose. The optimization course refines $\theta$ to physically realistic MANO parameters $\theta^\prime$ as the final result of synthesis.

To achieve better contact quality, we define a contact loss $\mathcal{L}_{\mathrm{contact}}$ to attract the nearby finger vertices to the local surface section. And to avoid penetration, we also use a penetration loss $\mathcal{L}_{\mathrm{penetr}}$ by repulsing hand mesh vertices that are inside the object mesh. 

Besides, three additional losses $\mathcal{L}_{\mathrm{trans}}(\theta^\prime_{\cdots;0,1,2})=\sum_{t}\lVert\theta^\prime_{t;0,1,2}-\theta^\prime_{t+1;0,1,2}\rVert^2$, $\mathcal{L}_{v}(\theta^\prime)=\lVert \dot{\mathbf{p}}\rVert^2$ and $\mathcal{L}_{a}(\theta^\prime)=\lVert \ddot{\mathbf{p}}\rVert^2$ are utilized for smoothening the wrist transition, velocity and acceleration of hand joints, respectively.

We minimize the overall loss value defined as

\vspace{-10pt}

\begin{equation} \label{eq:syn_all_loss}
\begin{split}
    \mathcal{L}(\theta^\prime) &= \lambda_{\mathrm{contact}}(\mathcal{L}_{\mathrm{contact}}(\theta^\prime)+\mathcal{L}_{\mathrm{penetr}}(\theta^\prime)) \\
    & + \lambda_{\mathrm{trans}}\mathcal{L}_{\mathrm{trans}}(\theta^\prime_{\cdots;0,1,2}) \\
    & + \lambda_{\mathrm{smooth}}(\lambda_{v}\mathcal{L}_{v}(\theta^\prime)+\lambda_{a}\mathcal{L}_{a}(\theta^\prime)).
\end{split}
\end{equation}

To produce more accurate contacts, we iteratively conduct such an optimization process for several epochs. In each epoch we feed the current $\theta^\prime$ to the optimization course for further improvement.

%% file: tex/exp.tex
\input{tex/tables/dataset.tex}

\begin{figure}
    \centering
    \vspace{-5pt}
    \includegraphics[width=0.9\linewidth]{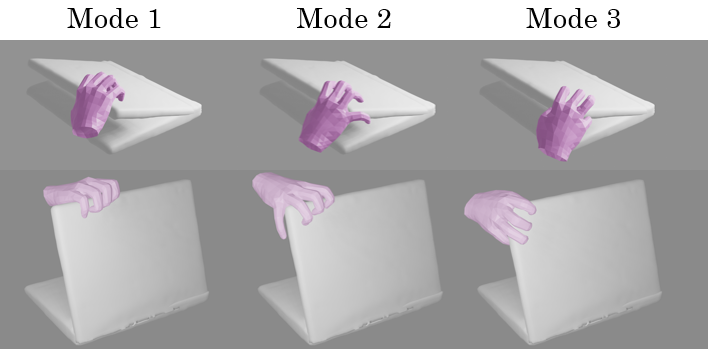}
    \caption{\textbf{Visualization of different grasp modes on the same laptop.} We show 3 different grasp modes for opening a laptop. Both the starting and ending hand poses of the grasps are shown. }
    \label{fig:result-modediversity}
\end{figure}

\begin{figure}
    \centering
    \vspace{-5pt}
    \includegraphics[width=0.9\linewidth]{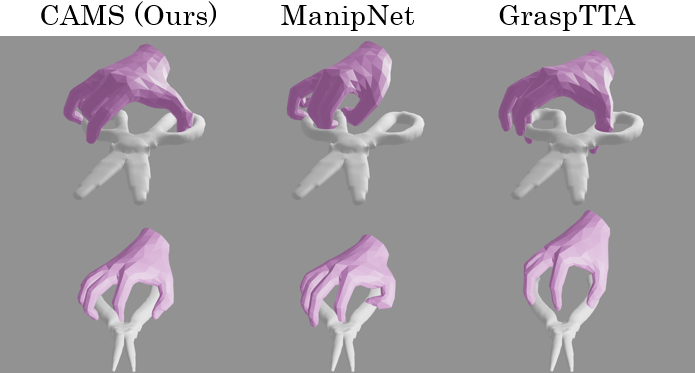}
    \caption{\textbf{Qualitative results compared with GraspTTA \cite{grasptta} and ManipNet \cite{zhang2021manipnet}.} The results generated by our method are more realistic.}
    \label{fig:result-comparison}
    \vspace{-10pt}
\end{figure}

\label{sec:exp}
In this section, we
apply our method to synthesize HOM, and evaluate the effect of our method with various metrics. We first introduce the experimental settings including dataset (Section~\ref{sec:data}), baselines (Section~\ref{sec:baselines}), and evaluation metrics (Section~\ref{sec:metrics}), while the experimental details are presented in our supplementary material.
We then show that our method could handle shape diversity and manipulation diversity and achieve a surpassing synthesis performance compared with other methods in Section~\ref{sec:compare}.

\subsection{Dataset}
\label{sec:data}

We utilize HOI4D Dataset \cite{hoi4d} in our experiment. HOI4D is a real-world dataset that contains dynamic HOM data spanning various rigid and articulated object categories. We select five object categories with different functionalities in our experiment, as shown in (\tabref{tab:dataset}). 
The selected HOM data contains both various manipulation types (\eg opening a laptop with different human preferences) and complex manipulation processes (\eg opening a small scissor by putting fingers through the hole) to improve the diversity and complexity of our synthesis results. To improve the usability of HOI4D under our setting, we applied some data cleaning and augmentation methods to the raw data (see our supplementary material for more details).

\input{tex/tables/evaluation.tex}

\input{tex/tables/userstudy.tex}

\subsection{Evaluation Metrics}
\label{sec:metrics}

We report several evaluation metrics on our experiments. We use these metrics to quantize the human-likeness and physical plausibility of motion synthesis results.

%

\textbf{Contact-Movement Consistency} We evaluate whether the object's movement can align with the contact forces produced by hand-object contacts, using the same physics model as in ManipNet\cite{zhang2021manipnet}. For articulated objects, we assume there are free opposite forces at the spin axis of the object. We calculate the proportion of frames that the contacts align with the object motion.

\textbf{Articulation Consistency} For articulated objects, we evaluate whether the hand pose can manipulate the object in a human-like manner. In particular, for each part of the object in each frame, we compute the torque of all contact points w.r.t. the object's spin axis if a unit force is applied along the normal direction of the contact point. If the maximal attitude of the torque with the same direction of object rotation exceeds a threshold, we regard such frame as qualified. We calculate the proportion of qualified frames.

\textbf{Penetration Rate} We compute the mean penetration proportion of hand vertices for each sequence. We regard penetrations within a small threshold $\lambda$=5mm as not penetrated since small penetrations can be seen as contacts made by a soft hand in real.

\textbf{Perceptual Score} We collect human perceptual scores to judge the naturalness of the motion sequences. We ask people not familiar with motion synthesis to give discrete perceptual scores for the results and calculate a mean score for each category using each baseline method. The detailed approach is left to the supplementary material.

\subsection{Baselines}
\label{sec:baselines}
As introduced in Section~\ref{sec:relatedwork}, there are only a few works about dynamic HOM generation using a learning-based method, while there are various techniques that could have the potential to be used in this task. Consequently, we design the following baselines.

\textbf{GraspTTA\cite{grasptta}:} GraspTTA proposed a strong baseline in static grasp generation. We use it to generate static grasps of several key snapshots in manipulation and then refine the result at test time using the TTA loss (We do not refine the network at test time). The dynamic HOM animation is thus generated from interpolation based on these snapshots.

\textbf{ManipNet\cite{zhang2021manipnet}:} Benefiting from carefully designed geometric sensors, ManipNet has shown a strong ability to generalization on generic object manipulation synthesis tasks. Different from our setting, ManipNet assumes additional input of wrist trajectory. To compare it with other methods, we provide it with the wrist trajectory generated by CAMS as input (and thus it only differs at fingers).

Though both GraspTTA and ManipNet have their own mechanisms to improve synthesis quality (\eg reduce penetration), they can also benefit from our optimization-based motion synthesizer (optimizing $\mathcal L_{penetr}$ to reduce penetration). We combine all baselines with an optimization stage (denoted as ``w/ opt'') and compare them with the CAMS-CVAE motion planner.

\subsection{Ablation Studies}
\label{sec:ablation}

\textbf{Remove Contact Optimization:} To demonstrate the advantages of contact optimizations in motion synthesis, we train a baseline model with the contact optimizations removed (CAMS-).

Besides removing the contact optimization in the synthesizer, we also did several ablation studies using different representation spaces of fingers. These experiments are left to our supplementary material.

\subsection{Comparison}
\label{sec:compare}

\textbf{Quantitative Results} Table~\ref{tab:quantitative_result} and Table~\ref{tab:user_study} show the quantitative results of our method and all baselines. Our method outperforms previous work on all tasks, even if contact optimization is applied to them.

An observation is that after applying the offline optimization stage, the performance gain of our method is significantly higher than the baselines. This can be explained by that the $\mathcal L_{contact}$ term in contact optimization takes the contact targets from our planner as input, and it is the key to producing gradients guiding the finger placement. Without intermediate contact target information, the optimization can only leverage the $\mathcal L_{penetr}$ loss term, and it may push the finger out of the object in unpredictable directions.

\textbf{Qualitative Results} Besides \figref{fig:Head}, \figref{fig:result-modediversity} shows that our method can generate diverse grasp modes on a single object instance. \figref{fig:result-comparison} shows that our method can generate reasonable poses given complex object shapes. We also show full result demonstrations in our video, including the generated whole HOM processes for different tasks, robustness to different object shapes and sizes, comparison between baseline methods, and diversity of generated manipulation styles.

%% file: tex/tables/dataset.tex
\begin{table}[t]
	\centering
	{
		\begin{tabular}{p{3cm} p{3cm}}
		
			\toprule

            Category & Task  \\
			
			\midrule
            Laptop & Open the Laptop \\
            Bucket & Lift the Handle \\
            Scissors & Open the Scissors \\
            Pilers & Clamp \\
            Kettle & Pick and Place \\
			
			\bottomrule
		\end{tabular}
	}
    \vspace{-0.1cm}
	\caption{\textbf{Categories and Tasks}. We select five object categories (four articulated object categories and one rigid object category) with different manipulation tasks from HOI4D \cite{hoi4d}.
	}
 \vspace{-15pt}
	\label{tab:dataset}
\end{table}{}

%% file: tex/tables/evaluation.tex
\begin{table*}[t]
\centering
\begin{tabular}{cccccccccc}
    \toprule[2pt]
    &\multicolumn{3}{c}{Pliers} & \multicolumn{3}{c}{Scissors}  & \multicolumn{3}{c}{Laptop}  \\ 
    & Pen $(\%)\downarrow$ & Mov $\uparrow$ & Art $\uparrow$ & Pen $(\%)\downarrow$ & Mov $\uparrow$ & Art $\uparrow$ & Pen $(\%)\downarrow$ & Mov $\uparrow$ & Art $\uparrow$  \\
    \midrule
    Ground Truth & 0.000 & 1.000 & 1.000 & 0.046 & 1.000 & 0.970 & 0.316 & 1.000 & 1.000 \\
    \midrule
    GraspTTA & 0.555 & 0.779 & 0.420 & 0.454 & 0.993 & 0.849 & 5.211 & \textbf{1.000} & 0.997 \\
    GraspTTA w/ opt & 0.294 & 0.727 & 0.321 & 0.812 & 0.994 & 0.959 & 4.702 & \textbf{1.000} & \textbf{1.000} \\
    ManipNet & 0.548 & 0.984 & 0.892 & 0.391 & 0.917 & 0.417 & 3.550 & \textbf{1.000} & 0.995 \\
    ManipNet w/ opt & 0.387 & 0.890 & 0.738 & 0.131 & 0.831 & 0.333 & 2.762 & \textbf{1.000} & 0.994 \\
    CAMS- & 0.563 & 0.916 & 0.393 & 0.590 & 0.997 & 0.850 & 5.204 & \textbf{1.000} & 0.983 \\
    CAMS (Ours) & \textbf{0.004} & \textbf{1.000} & \textbf{1.000} & \textbf{0.080} & \textbf{0.999} & \textbf{0.989} & \textbf{0.906} & \textbf{1.000} & \textbf{1.000} \\
    \midrule[2pt]
    & \multicolumn{3}{c}{Kettle}  &  \multicolumn{3}{c}{Bucket} &  \multicolumn{3}{c}{Overall} \\ 
    & Pen $(\%)\downarrow$ & Mov $\uparrow$ & Art $\uparrow$ & Pen $(\%)\downarrow$ & Mov $\uparrow$ & Art $\uparrow$ & Pen $(\%)\downarrow$ & Mov $\uparrow$ & Art $\uparrow$  \\
    \midrule
    Ground Truth & 0.602 & 1.000 & N/A & 0.090 & 1.000 & 1.000 & 0.211 & 1.000 & 0.992 \\
    \midrule
    GraspTTA & 4.852 & 0.586 & N/A & 1.309 & \textbf{1.000} & 0.863 & 2.476 & 0.872 & 0.782 \\
    GraspTTA w/ opt & 3.496 & 0.642 & N/A & 1.244 & \textbf{1.000} & 0.886 & 2.110 & 0.873 & 0.791 \\
    ManipNet & 0.760 & 0.892 & N/A & 0.231 & \textbf{1.000} & 0.844 & 1.096 & 0.959 & 0.787 \\
    ManipNet w/ opt & 0.459 & 0.622 & N/A & 0.156 & \textbf{1.000} & 0.865 & 0.779 & 0.869 & 0.733 \\
    CAMS- & 2.494 & 0.759 & N/A & 0.151 & \textbf{1.000} & 0.893 & 1.800 & 0.934 & 0.780 \\
    CAMS (Ours) & \textbf{0.098} & \textbf{0.915} & N/A & \textbf{0.015} & \textbf{1.000} & \textbf{1.000} & \textbf{0.221} & \textbf{0.983} & \textbf{0.997} \\
    \bottomrule[2pt]
    
\end{tabular}
\caption{\textbf{Quantitative results compared with GraspTTA \cite{grasptta} and ManipNet \cite{zhang2021manipnet}.} ``Pen'' denotes the average percentage of hand vertices penetrated in the object. ``Mov'' denotes the average proportion of frames that are contact-movement consistent. ``Art'' denotes the average proportion of frames that are articulation consistent (see Section \ref{sec:metrics}). }
\vspace{-0.5cm}
\label{tab:quantitative_result}
\end{table*}

%% file: tex/tables/userstudy.tex
\begin{table}[t]
\centering
\begin{tabular}{cccc}
    \toprule[2pt]
    & \multicolumn{3}{c}{Perceptual Score $\{1,\ldots,5\}\uparrow$} \\
    & Pliers & Scissors & Laptop \\
    \midrule
    Ground Truth & 4.145 & 4.025 & 3.980 \\
    \midrule
    GraspTTA & 1.243 & 2.550 & 1.450 \\
    ManipNet & 2.000 & 1.281 & 2.400 \\
    CAMS- & 1.986 & 1.200 & 1.980 \\
    CAMS (Ours) & \textbf{3.841} & \textbf{3.300} & \textbf{3.600} \\
    \midrule[2pt]
     & Kettle & Bucket & Overall \\
    \midrule
    Ground Truth  & 3.511 & 3.920 & 3.882 \\
    \midrule
    GraspTTA  & 2.146 & 1.520 & 1.717 \\
    ManipNet  & 1.988 & 2.143 & 2.069 \\
    CAMS-  & 2.202 & 2.860 & 2.101 \\
    CAMS (Ours) & \textbf{2.360} & \textbf{3.700} & \textbf{3.290} \\
    \bottomrule[2pt]
    
\end{tabular}
\caption{\textbf{Perceptual Score.}}
\vspace{-0.5cm}
\label{tab:user_study}
\end{table}

%% file: tex/conclusion.tex
In this work, we tackle a novel task of category-level functional hand-object manipulation synthesis. To generate human-like and physically realistic manipulation sequences, we design a two-level space hierarchy named CAnonicalized Manipulation Spaces (CAMS) and thus present a two-stage framework containing a planner and a synthesizer that leverage CAMS as an intermediate representation. Our method achieves state-of-the-art performance for both articulated and rigid object categories.
